\documentclass[letterpaper, 10 pt, conference]{ieeeconf}  

\IEEEoverridecommandlockouts                              

\usepackage{hyperref}
\hypersetup{bookmarksopen,bookmarksnumbered,
pdfpagemode=UseOutlines,
colorlinks=true,
linkcolor=blue,
anchorcolor=blue,
citecolor=blue,
filecolor=blue,
menucolor=blue,
urlcolor=blue
}
\usepackage{upgreek}
\usepackage{amsfonts,amssymb}
\usepackage{bm}
\usepackage{bbm}

\usepackage{amsthm} 
\usepackage{thmtools}
\usepackage{mathtools}
\usepackage{xspace}
\usepackage{units}
\usepackage{booktabs} 
\usepackage{tabulary}
\usepackage[usenames,dvipsnames,table]{xcolor} 
\usepackage{diagbox}
\usepackage[ruled,vlined,linesnumbered]{algorithm2e}  
\SetKwComment{Comment}{$\triangleright$\ }{}
\usepackage{ifthen,version}
\usepackage{soul}
\usepackage{csquotes}
\usepackage{microtype}      
\usepackage{lipsum}
\usepackage{tikz}
\let\labelindent\relax
\usepackage{todonotes}
\usepackage{enumitem}
\usepackage[noadjust]{cite}
\usepackage{wrapfig}
\usepackage{graphicx}
\usepackage{array}
\usepackage{subcaption}
\usepackage{nopageno}
\usepackage{cleveref}

\usepackage[normalem]{ulem}
\newcommand\redsout{\bgroup\markoverwith{\textcolor{red}{\rule[0.5ex]{2pt}{0.7pt}}}\ULon}

\newcommand{\xxnote}[3]{}
\ifx\hidenotes\undefined
  \usepackage{color}
  \renewcommand{\xxnote}[3]{\color{#2}{#1: #3}}
\fi

\usepackage{multirow}
\usepackage{pbox}

\newtheoremstyle{hypstyle}
{3pt} 
{3pt} 
{\itshape} 
{} 
{\bfseries} 
{.} 
{.5em} 
{} 

\theoremstyle{hypstyle}

\newcommand{\real}[0]{\mathbb{R}}

\newcommand{\bbm}{\begin{bmatrix}}
\newcommand{\ebm}{\end{bmatrix}}

\graphicspath{{figs/}}

\title{\LARGE \bf
LATTE: LAnguage Trajectory TransformEr
\vspace{-2mm} 

}

\author{Arthur Bucker$^{1}$, Luis Figueredo$^{1}$, Sami Haddadin$^{1}$, Ashish Kapoor$^{2}$, Shuang Ma$^{2}$, Sai Vemprala$^{2}$, Rogerio Bonatti$^{2}$\\[2mm]
$^{1}$Technische Universit{\"a}t M{\"u}nchen, $^{2}$Microsoft
\vspace{-3mm}
}

\begin{document}

\maketitle
\thispagestyle{plain}
\pagestyle{plain}



\begin{abstract}

Natural language is one of the most intuitive ways to express human intent. However, translating instructions and commands towards robotic motion generation and deployment in the real world is far from being an easy task. 
The challenge of combining a robot's inherent low-level geometric and kinodynamic constraints with a human's high-level semantic instructions 
traditionally is solved using task-specific solutions with little generalizability between hardware platforms, often with the use of static sets of target actions and commands. 
%
This work instead proposes a flexible language-based framework that allows a user to modify generic robotic trajectories.
Our method leverages pre-trained language models (BERT and CLIP) to encode the user's intent and target objects directly from a free-form text input and scene images, fuses geometrical features generated by a transformer encoder network, and finally outputs trajectories using a transformer decoder, without the need of priors related to the task or robot information.
We significantly extend our own previous work presented in \cite{bucker2022reshaping} by expanding the trajectory parametrization space to 3D and velocity as opposed to just XY movements. 
In addition, we now train the model to use actual images of the objects in the scene for context (as opposed to textual descriptions), and we evaluate the system in a diverse set of scenarios beyond manipulation, such as aerial and legged robots. 
Our simulated and real-life experiments demonstrate that our transformer model can successfully follow human intent, modifying the shape and speed of trajectories within multiple environments.
Codebase available at: \small{\url{https://github.com/arthurfenderbucker/LaTTe-Language-Trajectory-TransformEr.git}}.

\end{abstract}

\section{Introduction}
\label{sec:intro}
Robots are increasingly working in proximity to humans, sharing living and working spaces.
Within this context, it is of high importance for the robotics community to research techniques that allow autonomous agents to seamlessly interact with human users.
This work focuses on one important facet of human-robot interaction: given a user's objective and an obstacle environment, how can the robot best generate a trajectory that respects the human preferences while tending to safety and dynamics constraints in its surroundings?
%

\begin{figure}[t]
    \centering
    \includegraphics[width=0.45\textwidth]{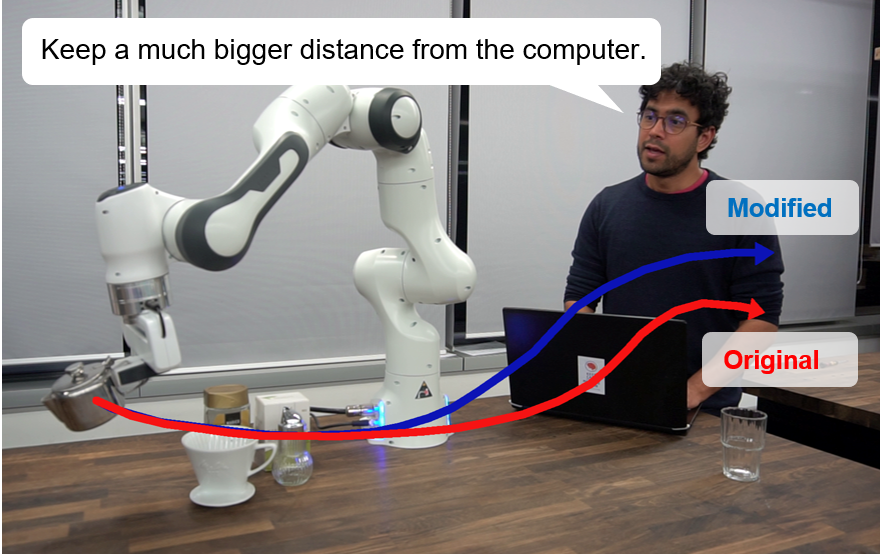}
    \caption{
    \small{Trajectory reshaping obeying user's constraints. Our method fuses natural language commands, images of the environment, and geometrical data to generate the modified robot's trajectory.}
    }
    \label{fig:main_fig}
    \vspace{-6mm}
\end{figure}

Robots of today are still largely pre-programmed for specific tasks, and have very limited capability to operate and adapt to new contexts among unstructured human-centered environments.
Ideally, in such scenarios the robot should have the ability to recognize and understand natural language commands in a given context and map them to the task-domain space
-- where tasks and constraints are largely influenced by context, intent and affordances with objects \cite{2015_Jain_Sharma_Joachims_Saxena_IJRR}.
This paradigm shift deviates from traditional motion planning, and requires methodologies that are able to integrate multi-modal inputs coming from perception systems (for instance user-provided language commands and robot vision) together with geometrical information to shape robot trajectories towards the desired human intent.
Figure~\ref{fig:main_fig} displays a typical application scenario for our trajectory adaptation method.

The core of our method lies within natural language understanding, which is the most intuitive way for a user to express their intent.
While large pre-trained large language models (LLMs) such as BERT~\cite{devlin2018bert}, GPT3~\cite{brown2020language} and Megatron-Turing~\cite{smith2022using} have revolutionized our ability to perform linguistic tasks in recent years, we have just started to see pioneering works that incorporate large foundation language models with robotics tasks \cite{bucker2022reshaping,gadre2022clip,shridhar2022cliport,ahn2022can,sharma2022correcting}.
The use of pre-trained LLMs is extremely beneficial within the robotics context because human-provided annotations are scarce and often costly to obtain.
The challenge which we explore in this paper then becomes how we can exploit these rich semantic representations and align them with geometrical trajectory data when mapping commands towards trajectory waypoints.



\begin{figure*}[t]
    \centering
    \includegraphics[width=0.87\textwidth]{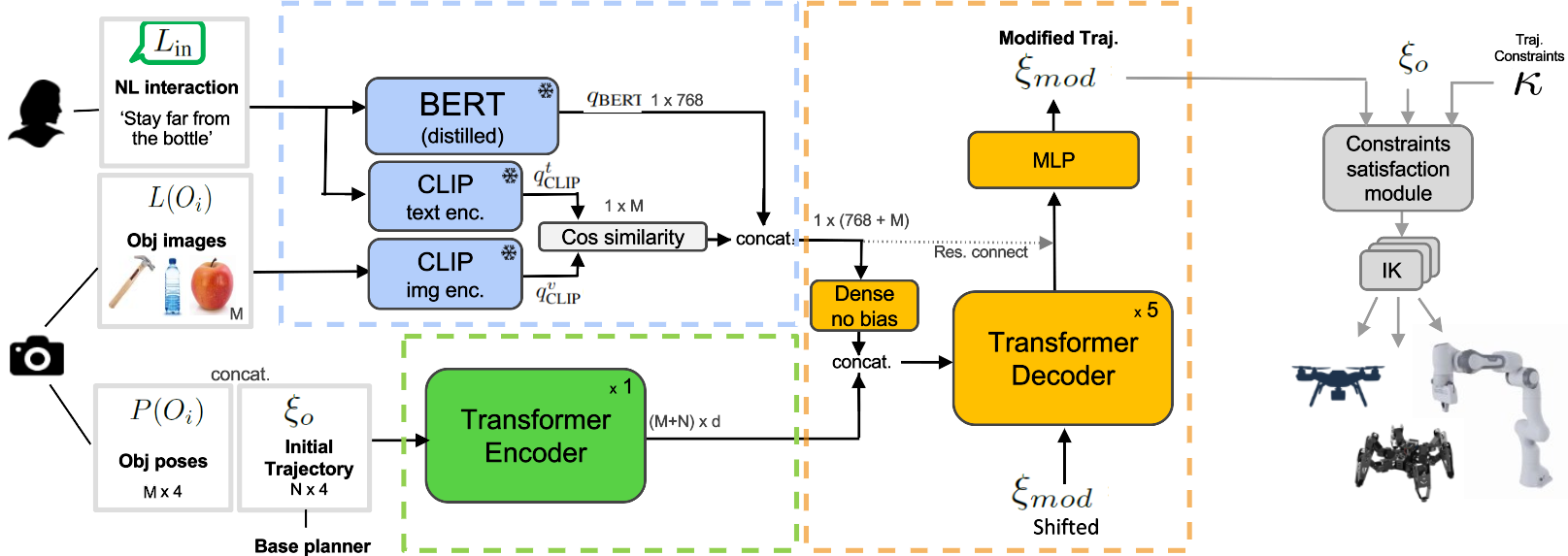}
    \caption{
    \small{Systems architecture: in blue, the language and contextual encoding module, compose mainly of frozen pre-trained models. In green the geometrical encoding . In orange the multimodal tranformer decoder.}
    }
    \label{fig:pipeline}
\end{figure*}


In this work we propose a framework that allows a user to reshape a trajectory using language instructions.
Our method uses a initialization from any geometrical planner (\textit{e.g.} A\textsuperscript{*}, RRT\textsuperscript{*} \cite{lavalle2006planning}, MPC~\cite{garcia1989model}), which are concerned solely about obstacle avoidance and dynamics constraints, and augments it with semantic objectives.
This paper serves as an extension of our previous work in this domain \cite{bucker2022reshaping}, but with significant improvements in the architecture and experimental evaluations:

\begin{itemize}
\item \textbf{Trajectory dimensionality:} we expand the dimension of each trajectory waypoint from planar (XY) to 3D and velocity in this work;
\item \textbf{Environment images:} while the original paper used textual object labels (\textit{e.g.} 'Hammer', 'Bottle') as input to the network, here we use images of objects when inferring targets for the user's commands, which is a more realistic setting;
\item \textbf{Multi-platform evaluation:} We expand the experimental evaluation towards multiple robotics form factors beyond manipulators. We show that the model's outputs are amenable to different robot dynamics and motion controller in aerial and legged locomotion domains.
\end{itemize}

\section{Related Work} 
\label{sec:related_work}

\textbf{Natural language and robotics:} 
Equipping robots with natural language models provides an intuitive and straightforward interface to address these challenges through human interaction and decision-making.
Classically, modeling human-robot interactions using language is challenging because it forces the user to operate within a rigid set of instructions~\cite{tellex2020robots}, or requires mathematically complex algorithms to keep track of multiple probability distributions over actions and target objects \cite{arkin2020multimodal,walter2021language}.
There has been an increase in recent works that explore the use of deep models to implicitly keep track of the complex mapping between language and actions, but the downside is that they often require vast amounts of data for training \cite{fu2019language,hong2011recurrent,stepputtis2020language,goyal2021zero}.

In the domain of navigation we find literature that investigates the use of multi-modal representations fusing natural language and perception along with planning modules through the use of cost functions or reinforcement learning \cite{huang2022language,shridhar2022cliport,hong2011recurrent,shao2021concept2robot,goodwin2021semantically,bommasani2021opportunities,szot2021habitat,Anderson2018room2room}.
In the manipulation domain we also find the work of \cite{shridhar2022cliport}, which uses CLIP~\cite{radford2021learning} embeddings to combine semantic and spatial information.
To this end, it can be often beneficial to use pre-trained multi-modal representations that align visual and language inputs representation such as \cite{sun2019videobert, lu2019vilbert, zhou2020unified, su2019vlbert}, which often using BERT-style \cite{devlin2018bert} training procedures. 
Representations are often fine-tuned \cite{hao2020genericVLN,thomason2020vision,nguyen2019helpAnna} on the deployment scenario.

\textbf{Transformers for robotics:}
Transformers, originally introduced in the language processing ~\cite{vaswani2017attention}, quickly proved to be useful in modeling long-range data dependencies other domains.
Within the robotics motion planning context, transformers architectures have been directly used for trajectory forecasting~\cite{giuliari2021transformer} and reinforcement learning~\cite{chen2021decision,janner2021offline}.
A more common use of transformers in robotics has been as feature extraction modules for one or more modalities simultaneously that leverage large-scale pre-trained models \cite{bucker2022reshaping,ahn2022can,gadre2022clip,shridhar2022cliport,sharma2022correcting}.

Particularly close to this paper is the work of \cite{sharma2022correcting}. It uses pre-trained LLMs to create a semantic cost map that guides a optimization-based motion planner to produce trajectories that satisfy motion constraints provided by a user in free-form text.
Similarly, our method also uses LLMs for textual and visual feature extraction, however we use a transformer encoder-decoder pair to align semantic information with geometric cues to recast trajectories.
Our first paper presented in \cite{bucker2022reshaping} validated our approach for 2D scenarios, and showed its effectiveness compared to other interfaces for human-robot interaction.
As described at the end of section~\ref{sec:intro} this paper extends these ideas to higher dimensions and more realistic experimental settings.

\section{Approach}
\label{sec:approach}

Our overall goal is to provide a flexible interface for human-robot interaction within the context of trajectory reshaping that is agnostic to robotic platforms. 
The user provides a natural language command, and the robot's body or end-effector behavior, which is expressed with a 3D trajectory over time, is expected to be modified accordingly. 
Our trajectory generation system uses a sequential waypoint prediction model that takes into account multiple data modalities from scene geometry, environment images and the language input, all of which are fed into a transformer encoder-decoder pair.

Beyond the user's semantic intent, we expect the final trajectory to also respect safety and dynamics space-state constraints, which can be achieved by post-processing the model's output into a continuous state space.
This last stage allows our same model to be employed by different robot form factors by using the proper inverse kinematics modules.


\subsection{Problem Definition}

Let $\xi_{o}: [-1,1] \rightarrow \mathbb{R}^4$ be the original normalized robot trajectory 
which is composed by a collection of $N$ waypoints and associated velocities $\xi_{o} = \{(x_1,y_1,z_1, v_1),...,(x_N,y_N,z_N, v_N))\}$, where $x_i, y_i, z_i$ and $v_i$ are the waypoint coordinates and the velocity at time step $i$, respectively.
We assume that the original trajectory 
obeys the system constraints 
and can be pre-calculated using any desired motion planning algorithm, but falls short of the full task specifications.
Let $L_{\text{in}}$ be the user's natural language input sent to correct the original trajectory, such as $L_{\text{in}}=\text{``Go slower when next to the fragile glasses''}$.


Let $\mathcal{O}=\{O_1, ..., O_M\}$ be a collection of $M$ objects in the environment, each with a corresponding position $P(O_i) \in \mathbb{R}^3$ and image $I(O_i)$. Our goal is to learn a function $f$ that maps the original trajectory, user command and obstacles towards a modified trajectory $\xi_{mod}$, which obeys the user's semantic objectives and is contained in the system feasible domain $K$:

\vspace{-2mm}
\begin{equation}
    \label{eq:main}
    \xi_{mod} = f(\xi_{o}, L_{\text{in}}, \mathcal{O})
\end{equation}

\subsection{Proposed Network Architecture}
\label{subsec:achitecture}



%
%

We approximate function $f$ from \eqref{eq:main} by a parametrized model $f_{\theta}$, learned directly in a data-driven manner. 
This mapping is non-trivial since it combines data from multiple distinct modalities, and also contains ambiguities in solution space since there are multiple trajectories that satisfy the user's semantic objective.

Our model architecture is divided into 3 main modules and one constraint satisfaction step. Fig.~\ref{fig:pipeline} shows the connection between this modules.
First, a language and image encoder makes use of distinct pre-trained feature encoders (BERT and CLIP) to generate a embedded representation of the natural language input and to identify the possible objects referred to in the text.
Next, a geometry encoder uses object poses and trajectory waypoints as inputs and uses a transformer to learn geometric relations between the original trajectory, speed profiles and the objects in the scene. 
Finally, a multi-modal transformer decoder combines the embedded outputs of the two prior modules to generate the modified trajectory autoregressively. 
We discuss each module in detail below:


\textbf{Language and image encoder:}
The use of a large language model creates more flexibility in the natural language interface, allowing the use of synonyms (shown in Section~\ref{subsec:sim_exp}) and less training data, given that the encoder has already been trained with a massive corpus. 
We use a pre-trained BERT encoder~\cite{devlin2018bert}, to produce semantic feature 
$q_\text{BERT}(z | L_{\text{in}})$ from the user's input. 
In addition, we use the pre-trained text and image encoders from CLIP~\cite{radford2021learning} to extract latent embeddings from both the user's text $q^t_\text{CLIP}(z|L_{in})$ and the $M$ object images $q^v_\text{CLIP}(z|I(O))$. 
We compute the cosine similarity vector $s$ between the visual and textual embeddings in order to identify a possible target object for the user's command. 
In section \ref{subsec:sim_exp} we show that using the object's images for target identification brings equivalent results as our previous work \cite{bucker2022reshaping} with object textual descriptions, since CLIP maps both modalities to a joint latent space.
Finally, we concatenate the similarity vector $s$ and the semantic features $q_\text{BERT}(z | L_{\text{in}})$ forming what we call semantic embedding $q_\text{S}$.


\textbf{Geometry encoder:}
The original trajectory $\xi_{o}$ is composed of points that are low-dimensional tuples $(x_i,y_i,z_i,v_i) \in \mathbb{R}^4$. 
In order to extract more meaningful information from each waypoint, we follow the example of \cite{giuliari2021transformer} and apply a linear transform with learnable weights $W_\text{geo}$ that projects each of these points into a higher dimensional feature space.
The poses $P(O_i)$ of each object are also processed with the same linear transform, and padded with zeros for the velocity component.

We then concatenate the sequences of high-dimensional feature vectors from waypoints and objects and use a transformer-based feature encoder $\text{T}_\text{enc}$ to extract geometrical features for each element.
The use of a Transformer model is preferred for sequences over recurrent networks because its architecture can intrinsically attend to multiple time steps simultaneously.
Conversely, recurrent networks suffer from vanishing gradient issues~\cite{giuliari2021transformer}, which negatively affect feature extraction and training stability.

\textbf{Multi-modal transformer decoder:}
Feature embeddings from both language and geometry are combined as input to a multi-modal transformer decoder block $\text{T}_\text{dec}$. 
This block generates the reshaped trajectory $\xi_{mod}$ sequentially, feeding the last token prediction as input to the next waypoint prediction.
This procedure is analogous to common transformer-based approaches for language translation~\cite{brown2020language,vaswani2017attention}, but in this case we can image that our model \textit{translates} trajectories from the original feature space towards a new space that obeys the user's semantic constraints.
We use imitation learning to train the model, and employ the Huber loss~\cite{huber1992robust} between the predicted and ground-truth waypoints.

\begin{figure*}[t]
  \centering
    \includegraphics[width=1.0\textwidth]{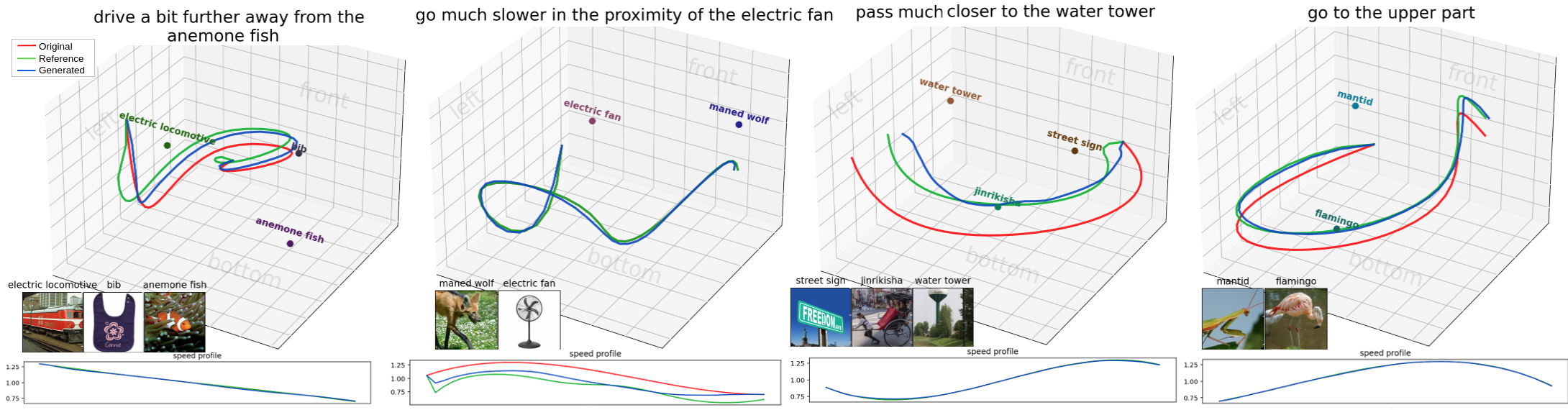}
  \caption{\small{Procedural dataset examples showing the original trajectory (red), ground-truth modifications, and model predictions (blue). Images representing objects are crawled from the web (bottom left), and the speed profile can also be modified (bottom right).}}
  \label{fig:random_exp} 
  \vspace{-5mm}
\end{figure*}

\subsection{Post-processing and execution}
\label{subsec:postprocessing}

Once a trajectory is generated by our model it needs to be post-processed to allow for the robot's execution. 
The modules described here allow our method to be agnostic to specific robotics platforms.

\textbf{Constraint satisfaction:} 
Constraint satisfaction is a complex and open field of study in robotics.
In this work we establish two simplifying assumptions regarding our deployment objectives.
First, the base motion planner outputs a set of hard constraints $K$ defined in the Cartesian space that define an admissible region for the trajectories. 
Second, we assume that the original trajectory is already within in the allowable constraint set.
We post-process our model's output trajectory by taking steps starting at the original waypoint towards the direction of new one: 
$\xi(t) = \xi_{o}(t) + \alpha (\xi_{mod}(t) - \xi_{o}(t))$, where $0<\alpha\leq1$.
If at any step we find that one waypoint reaches an inadmissible region then its position is not further updated.
We note that more complex constraint satisfaction algorithms can be developed here, but the simple approached described worked well with our scenarios.

\textbf{Inverse kinematics:}
Once the final trajectory is obtained, the user may plug in any inverse kinematics algorithm to obtain final trajectories for higher-dimensional degree of freedom robots.
In this work we evaluate our system with manipulators, aerial and legged robots.

\subsection{Synthetic Data Generation}
Data collection in the robotics domain can be challenging and expensive, specially when we require alignment between multiple modalities such as language, vision, and geometry.
We find different strategies in the robotics literature to deal with these issues, ranging from costly large-scale online user studies for language labeling~\cite{bonatti2021batteries,mandlekar2018roboturk} all the way to procedural data generation using heuristics~\cite{stepputtis2020language}.
Our work relies on purely procedural generation of trajectory-language pairs. 
We make a key hypothesis that the use of large-scale language models for feature encoding ($q_\text{BERT}$, $q_\text{CLIP}$) reduces the data requirements in terms of vocabulary diversity. 
We assume that if we are able to procedurally generate a small but meaningful set of examples with semantically-driven trajectory modifications we can train an effective transformer decoder, given that the BERT and CLIP encoders have already been trained with large corpuses and are able to handle vocabulary and sentence variations.
These assumptions are validated experimentally in Section \ref{sec:userstudy}.

Each data sample is composed of a base trajectory $\xi_0$, a natural language input $L_{in}$, a modified trajectory $\xi_{mod}$, and a set of object $\mathcal{O}=\{O_1, ..., O_M\}$ represented as central poses $P(O)$ and images $I(O)$.
$\xi_0$ is generated by fitting a spline in the Cartesian space through points generated in a random walk.
Objects poses are then randomly generated in space, and we sample object names from the Imagenet dataset \cite{deng2009imagenet} as their labels, and obtain various images for each one using a crawler over Bing Images using the object name as the web query.

As for the language input $L_{in}$, we focus on three main trajectory modifications: i) changes in the absolute Cartesian trajectory space (\textit{e.g.} ``stay on the left", ``go more to the right"), ii) changes in speed (e.g. ``go faster", ``go slower when next to \textit{x}"), and iii) positional changes relative to objects (e.g. ``walk closer to \textit{x}", ``drive further away from \textit{x}").
We pick a sample from a vocabulary bank associate each modification type, and calculate a force vector field over the enviornment using a handcrafted function $F(L_{in}, P(O))$. The field strength may vary depending on additional intensifier words that can be added to the sentences such as ``very", ``a bit", etc. 
In the section \ref{subsec:sim_exp} we also explore augmenting these language inputs using BART \cite{lewis2019bart}, which is a pre-trained paraphrasing model.
Finally, we generate the ground-truth trajectory modification by iteratively optimizing the original trajectory along the vector field. 

We introduce one additional hyper-parameter in the dataset generation and model training which we name \textit{locality factor}.
For the same language prompt, some robotics contexts might require small localized trajectory changes while others might expect long-range modifications. 
After training, the locality factor allows the user to define their desired range of model influence.

\section{Experiments} 
\label{sec:results}
We conducted several simulated and real-world experiments to validate our methods.
Our main goals were to: i) measure the effectiveness of our trajectory modification algorithm in 3D and velocity space, ii) understand the influence of the different architectural components towards the model's success, and iii) validate the applicability of the model to multiple robotic platforms.

\subsection{Model training details}

We trained and evaluated the model described in \Cref{sec:approach} over a dataset containing $100$k examples of procedurally generated trajectory modification. Among these, we used $70$k samples for training, $10$k for validation and $20$k for testing.
We kept both BERT and CLIP encoder weights frozen in other to avoid biasing the models towards our vocabulary, with $q_\text{BERT}(z | L_{\text{in}}) \in \real^{768}$ and $q^v_\text{CLIP}(z|I(O)) \in \real^{512}$.
We upscale the dimensionality of each scene object pose from $4 \to 400$ (depth) using a learned linear matrix, and apply the same procedure to $40$ waypoints from the original trajectory $\xi_{o}$.
$\text{T}_\text{enc}$ is a 1-block transformer encoder, and $\text{T}_\text{dec}$ is a 5-block transformer.
Each transformer has 3 hidden layers with $512$ fully-connected neurons with Relu activations,one Layer Normalization, 8 attention heads.
We use the AdamW~\cite{you2019large} optimizer with an initial learning rate $\gamma = 1e-4$, a linear warm-up period of $15$ epochs and a learning rate decay of 10\% after a plateau of 10 epochs on the validation loss.
We use a Nvidia Tesla V100 GPU with batch size of $16$, and train the model for $500$ epochs in approximately $2$ hours.

\begin{figure}[t]
    \begin{center}
    \includegraphics[width=0.44\textwidth]{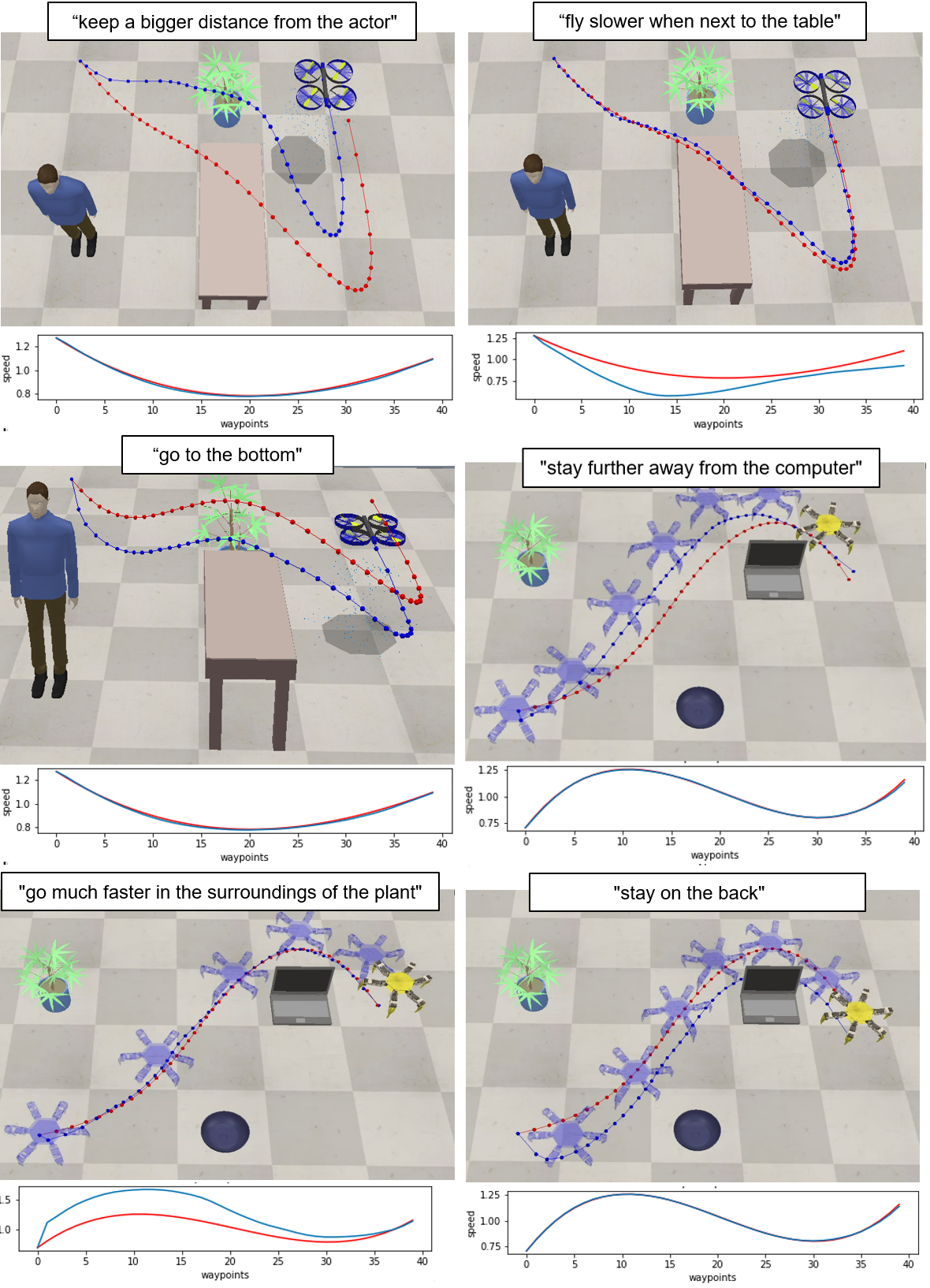}
    \end{center}
    \vspace{-4mm}
    \caption{\small{Model deployed with different robot form factors (drone and legged hexapod) for obstacle avoidance, speed refinement and absolute cartesian changes. Original trajectory shown in red, modification in blue, and corresponding speed profiles below each scenario. }}
    \label{fig:form_factors}
    \vspace{-6mm}
\end{figure}

\subsection{Simulation Experiments}
\label{subsec:sim_exp}

We apply our method to several simulated scenarios. First, we show the basic workings of our trajectory adaptation method through qualitative results which can be visualized in Figure \ref{fig:random_exp}. In this scenario, we use sample objects that were randomly chosen from crawling the web and their corresponding images. Assuming there is an initial trajectory that traverses around these objects, and given language commands indicating how to modify the trajectory (farther/closer to the object, faster/slower in the vicinity of an object), our model predicts trajectories that account for user intent. We show both spatial modifications as well as changes in speed profile in the trajectories output by our model.

\textbf{Multi-platform evaluation:}
To validate our framework's ability to adapt to different robot dynamics and environments we designed simulated environments using the CoppeliaSim simulator with Bullet physical engine \cite{coppeliaSim}.
While our original training dataset presents itself in a format amenable to end-effector positions within a manipulation context, this new simulator allows us to test our system on distinct robotic platforms, dynamics and base motion controllers. 

Specifically, we employ an aerial vehicle and a legged hexapod platform. 
The drone operates within a 3D global frame of reference and uses PID motion controller for trajectory tracking. 
In contrast, the hexapod is constrained to 2D movements and uses an open-loop motion controller.
As figure \ref{fig:form_factors} shows, our approach can successfully modify the base trajectories (red) for different types of natural language inputs. Additional experiments can be seen in the video attachment.

\textbf{Baseline architectures:}
We compare our proposed multi-modal transformer against architecture variations.
Table~\ref{tab:baselines} shows the result of a grid search over the number of layers and encoding dimension (depth) of the transformer encoder and decoders.
The model with one encode layer, 5 decoder layer and an depth of $400$ was chosen to be the reference model for our architecture and further baseline comparisons. 
We measure performance in terms the similarity between our model's output and the ground-truth trajectory modification in the dataset. 
Our metrics are MSE (mean squared error), MAE (mean absolute error), DTW (dynamic time warping), and DFD (discrete Frechet distance).

\begin{table}[tbh]
    \centering   
    \begin{tabular}{c|c|c|c||c|c|c|c}
    \toprule
         \kern-0.5emn.enc\kern-0.5em&\kern-0.5emn.dec\kern-0.5em&\kern-0.5emn.depth\kern-0.5em&\kern-0.5emparam.\kern-0.5em&MSE$\downarrow$ &  MAE$\downarrow$ &DTW$\downarrow$ & DFD$\downarrow$ \\ \hline
        
        2 & 3 & 256 & 4.95M & 0.00306 & 0.0314 & 3.1085 & 0.1346\\
        2 & 3 & 400 & 9.28M & 0.00235 & 0.0273 & 2.6966 & 0.1198\\
        2 & 5 & 256 & 6.53M & 0.00280 & 0.0284 & 2.8455 & 0.1265\\
        2 & 5 & 400 & 12.7M& 0.00238 & 0.0231 & 2.4900 & 0.1152\\
        1 & 3 & 256 & 4.42M& 0.00274 & 0.0272 & 2.8122 & 0.1245\\
        1 & 3 & 400 & 8.22M& \textbf{0.00224} & \textbf{0.0229} & \textbf{2.4445} & \textbf{0.1130}\\
        1 & 5 & 256 & 6.00M& 0.00277 & 0.0264 & 2.7527 & 0.1238\\
        1 & 5 & 400 & 11.2M & 0.00234 & 0.0227 & 2.4699 & 0.1138\\

        
         \bottomrule
    \end{tabular}
    \caption{\small{Architecture variations}}
    \label{tab:baselines}
    \vspace{-4mm}
\end{table}

Table~\ref{tab:baselines} provides valuable findings regarding the model architecture. For instance, increasing the number of encoder blocks caused no improvement on the model's performance. Furthermore the model with 3 decoder blocks presented slightly better results than the assumed baseline of 5 decoder block. 

In addition to model size, in table~\ref{tab:baselines2} we compare different architecture structures. The \textit{Naive} approach simply copies  the original trajectory. The \textit{No NL input} baseline represents a universal prior of the dataset, with an empty language command. \textit{Ours light} is a more compact version of our  model with 1 enc., 3 dec. and depth of 256.

\begin{table}[h]
    \centering
     \begin{tabular}{l|c||c|c|c|c}
    \toprule
         Approach & Param. & MSE$\downarrow$ &  MAE$\downarrow$ &DTW$\downarrow$ & DFD$\downarrow$ \\ \hline
        
        Naive & - & 0.00437 & 0.02709 & 3.568 & 0.1387\\
        No NL input & 11.2M & 0.04193 & 0.1663 & 15.097 & 0.5674 \\
        Ours light & 4,42M& 0,00274 & 0,0272 & 2,8122 & 0,1245\\
        Ours & 11.2M &  \textbf{0,00234} &\textbf{ 0,02273} & \textbf{2,4699} & \textbf{0,1138}\\
        
         
         
         \bottomrule
    \end{tabular}
    \caption{\small{Baseline architecture comparisons}}
    \label{tab:baselines2}
    \vspace{-4mm}
\end{table}

\begin{wrapfigure}{R}{0.25\textwidth}
\vspace{-6mm}
    \begin{center}
    \includegraphics[width=0.25\textwidth]{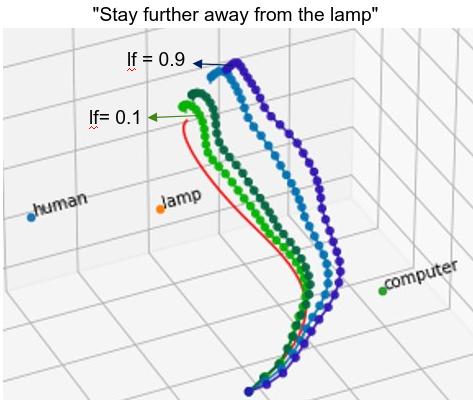}
    \end{center}
    \vspace{-4mm}
    \caption{\small{Locality factor influence}}
    \label{fig:locality_factor}
    \vspace{-3mm}
\end{wrapfigure} 

\textbf{Locality factor:}
Fig.~\ref{fig:locality_factor} shows the response of our model for different values of the locality factor (LF).
This hyper-parameter provides useful information on the range of the desired change change over the trajectory, which can serve as a finer user control besides the language input itself.


\textbf{Dataset size and augmentations:}
Table \ref{tab:textua_aug} shows the effect of increasing the training dataset size in model performance, as well as the effect of applying augmentations in the training data.
An increase in the dataset size from $1$k to $10$k samples significantly improves the validation metrics with minimal challenges besides a longer training time, given that data can be generated procedurally without expensive human annotations.
The geometrical augmentation (randomly shifting and scaling operations) shows a modest increase in performance. 

\begin{table}[htb]
\vspace{-2mm}
    \centering
    \begin{tabular}{l|c|c|c|c}
    \toprule
        & \multicolumn{4}{c}{Without geometrical augmentation}\\

         Dataset size & MSE$\downarrow$ &  MAE$\downarrow$ &DTW$\downarrow$ & DFD$\downarrow$ \\ \hline
         1k & 0.02608 & 0.11063 & 8.20700 & 0.46488 \\
        10k & 0.00243 & 0.02347 & 2.47016 & 0.11683 \\
        100k & 0.00229 & 0.02201 & 2.39301 & 0.11175 \\

        
         \hline
         & \multicolumn{4}{c}{With geometrical augmentation} \\
         Dataset size &  MSE$\downarrow$ & MAE$\downarrow$ &  DTW$\downarrow$ & DFD$\downarrow$ \\ \hline
        1k & 0.01420 & 0.07590 & 5.35290 & 0.35737 \\
        10k & 0.00248 & 0.02324 & 2.50841 & 0.11593 \\
        100k & 0.00234 & 0.02273 & 2.46992 & 0.11383 \\
        

    \bottomrule
    \end{tabular}
    \caption{
    \small{
    Effect of dataset size and geometrical augmentation. 
    }}
    \label{tab:textua_aug}
    \vspace{-2mm}
\end{table}

\subsection{Real Robot Experiments with Manipulation}

\begin{figure}[h]
    \centering
    \includegraphics[width=0.42\textwidth]{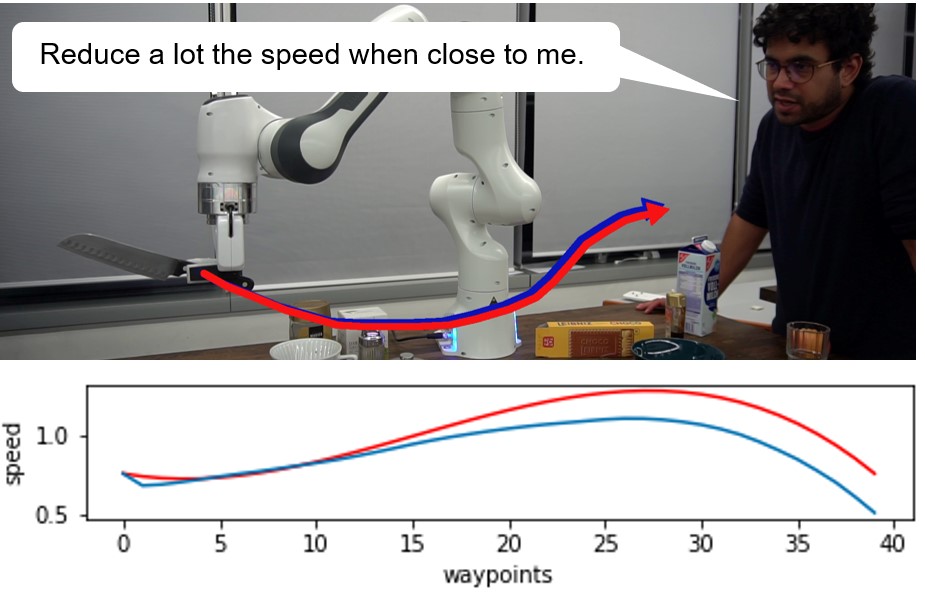}
    \caption{\small{Real life setup and sample interaction. It depicts the speed modification through online language instructions. An approximated representation of the original trajectory is shown in red, while the modified one in blue. Full videos in the supplementary material.}}
    \label{fig:realrobot}
    \vspace{-5mm}
\end{figure}

We deployed our model in real-world experiments using a 7-DOF PANDA Arm robot equipped with a claw gripper.
An off-the-shelf CPU/GPU setup computes the arm's low-level controller and our model. 
A camera mounted on the workbench captures images of the obstacle setting, and a YOLOV3\cite{redmon2018yolov3} object detector extracts bounding boxes of the five most likely objects to be sent to the CLIP encoder.
Snapshots of the setup and results can be found in figures \ref{fig:main_fig} and \ref{fig:realrobot}.
Additional experiments shown in the video attachment.

\subsection{User study experiments}
\label{sec:userstudy}
\begin{figure}[h]
    \centering
    \includegraphics[width=0.50\textwidth]{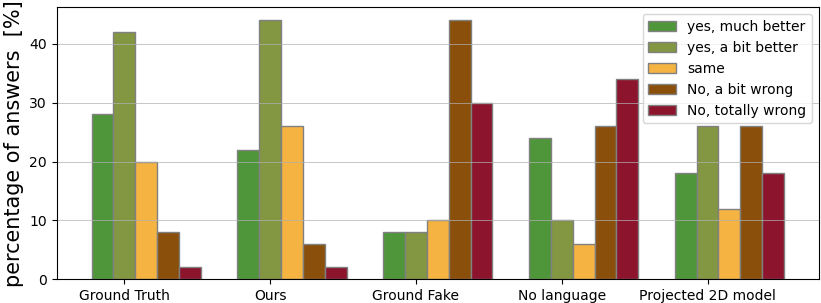}
    \caption{\small{Userstudy distributions of answer for each baseline.}}
    \label{fig:userstudy}
    \vspace{-2mm}
\end{figure}

We evaluated the model's performance against baseline architectures in a user study, collecting in total 300 data-points from 10 participants. Each user was asked to evaluate within a 1-5 Likert scale the trajectory changes generated from 5 different approaches considering a given NL interaction. Figure \ref{fig:userstudy} summarizes the distribution of answers for each baseline. \textit{"Ground Truth"} represents the procedural dataset used for training. As the chart shows, most users considered that our trajectory modifications in the dataset correctly represented the language commands. 
A similar pattern emerged from our trained model (\textit{"Ours"}), which yielded high-quality ratings.
The \textit{"Ground Fake"} approach shows samples of the dataset with intentionally wrong modifications, opposite to the ground truth, for the means of comparison. Non surprisingly it is rated with the lowest score.
The \textit{"No language"} baseline was also badly evaluated, showing that the model's performance is highly dependent on the language input, and that the model does not memorizes bias purely based on the scene context. 
Finally, the \textit{"Projected 2D"} distribution shows a direct comparison with our previous work \cite{bucker2022reshaping}, which produces pure 2D trajectory modifications. Its bad performance motivates the importance of the additions of 3D and velocity space that we incorporate in this paper.

\vspace{-2mm}

After the initial evaluations, each user was asked to freely interact with $5$ trajectories using a text box, and next judge the quality of the generated modifications. $48$\% of the user inputs presented words never seen by the model during the training process (out of distribution).
Even under these challenging conditions the model only failed on 24\% of the cases.
Table \ref{tab:out_of_dist} compares our model's performance for in and out of distribution settings.

\begin{table}[htb]
\vspace{-2mm}
    \centering
    \begin{tabular}{l|c|c|c}
    \toprule

         Textual interaction & Better [\%] & Same [\%] & Worse [\%] \\ \hline
         In-dataset vocabulary & 66.0  & 26.0 & 8.0\\
         Free user input & 46.0 & 30.0 & 24.0\\
    \bottomrule
    \end{tabular}
    \caption{
    \small{
    Evaluation of out of distribution NL interactions
    }}
    \label{tab:out_of_dist}
    \vspace{-2mm}
\end{table}

\vspace{-6pt}
\section{Conclusion and Discussion}
\label{sec:discussion}

This work develops a flexible language-based human-robot interface that
allows a user to modify existing robotic trajectories. Our method
leverages pre-trained large language and image models (BERT and CLIP) to
encode the user’s intent and target objects directly from a free-
form text input and scene images, fuses geometrical features generated by a transformer encoder network, and outputs
trajectories using a transformer decoder.

Our model can operate manipulate robot trajectories in 3D and velocity spaces.
The output trajectory can be post-processed and applied towards diverse
different platforms such as manipulation, aerial vehicles and legged robots. 
We provide a comprehensive set of simulated and real-world experiments demonstrating the effectiveness of our model and highlighting insights into what the model is learning.

In future iterations of this work we seek to explore additional modalities such as force inputs, as well as the ability of the model to interact with the user over longer time horizons and multiple instruction inputs. 
We hope that our framework can serve as a building block for a novel paradigms in human-robot collaboration that employ large language models. 


%




\footnotesize{
\bibliographystyle{IEEEtran}
\bibliography{IEEEexample.bib}
}


\end{document}